\documentclass{article}

\usepackage{microtype}
\usepackage{graphicx}
\usepackage{subcaption}
\usepackage{booktabs} 
\usepackage{tabularx} 
\usepackage{multirow} 
\usepackage{afterpage} 

\usepackage{hyperref}
\usepackage{bookmark}

\usepackage[preprint]{icml2026}

\usepackage{amsmath}
\usepackage{amssymb}
\usepackage{mathtools}
\usepackage{amsthm}
\usepackage{listings}
\usepackage{xcolor}

\definecolor{promptbg}{RGB}{248,248,248}
\definecolor{promptframe}{RGB}{200,200,200}
\lstdefinestyle{prompt}{
  backgroundcolor=\color{promptbg},
  frame=single,
  rulecolor=\color{promptframe},
  basicstyle=\ttfamily\footnotesize,
  breaklines=true,
  breakatwhitespace=true,
  columns=fullflexible,
  keepspaces=true,
  aboveskip=1em,
  belowskip=1em,
  xleftmargin=0.5em,
  xrightmargin=0.5em,
  framexleftmargin=0.3em,
  framexrightmargin=0.3em,
}

\usepackage[capitalize,noabbrev]{cleveref}

\theoremstyle{plain}

\theoremstyle{definition}

\theoremstyle{remark}

\usepackage[textsize=tiny]{todonotes}

\icmltitlerunning{Submission and Formatting Instructions for ICML 2026}

\begin{document}

\twocolumn[
  \icmltitle{DALDALL: Data Augmentation for Lexical and Semantic Diverse
  in Legal Domain by leveraging LLM-Persona}



  \icmlsetsymbol{equal}{*}

  \begin{icmlauthorlist}
    \icmlauthor{Janghyeok Choi\thanks{Undergraduate student.}}{snu,equal}
    \icmlauthor{Jaewon Lee\thanks{Ph.D.\ candidate.}}{snu,equal}
    \icmlauthor{Sungzoon Cho\thanks{Professor.}}{snu}
  \end{icmlauthorlist}

  \icmlaffiliation{snu}{Department of Industrial Engineering, Seoul National University, Seoul, South Korea}

  \icmlcorrespondingauthor{Sungzoon Cho}{zoon@snu.ac.kr}

  \icmlkeywords{Data Augmentation, Legal Domain, LLM-Persona}

  \vskip 0.3in
]



\printAffiliationsAndNotice{\icmlEqualContribution}

\begin{abstract}
  Data scarcity remains a persistent challenge in low-resource domains.
  While existing data augmentation methods leverage the generative capabilities
  of large language models (LLMs) to produce large volumes of synthetic data,
  these approaches often prioritize quantity over quality and lack domain-specific
  strategies. In this work, we introduce DALDALL, a persona-based data augmentation
  framework tailored for legal information retrieval (IR). Our method employs
  domain-specific professional personas—such as attorneys, prosecutors, and
  judges—to generate synthetic queries that exhibit substantially greater lexical
  and semantic diversity than vanilla prompting approaches. Experiments on the
  CLERC and COLIEE benchmarks demonstrate that persona-based augmentation achieves
  improvement in lexical diversity as measured by Self-BLEU scores, while
  preserving semantic fidelity to the original queries. Furthermore, dense retrievers
  fine-tuned on persona-augmented data consistently achieve competitive or superior
  recall performance compared to those trained on original data or generic augmentations.
  These findings establish persona-based prompting as an effective strategy for generating
  high-quality training data in specialized, low-resource domains.
\end{abstract}

\section{Introduction}

\begin{figure*}[t]
      \centering
      \includegraphics[width=\textwidth]{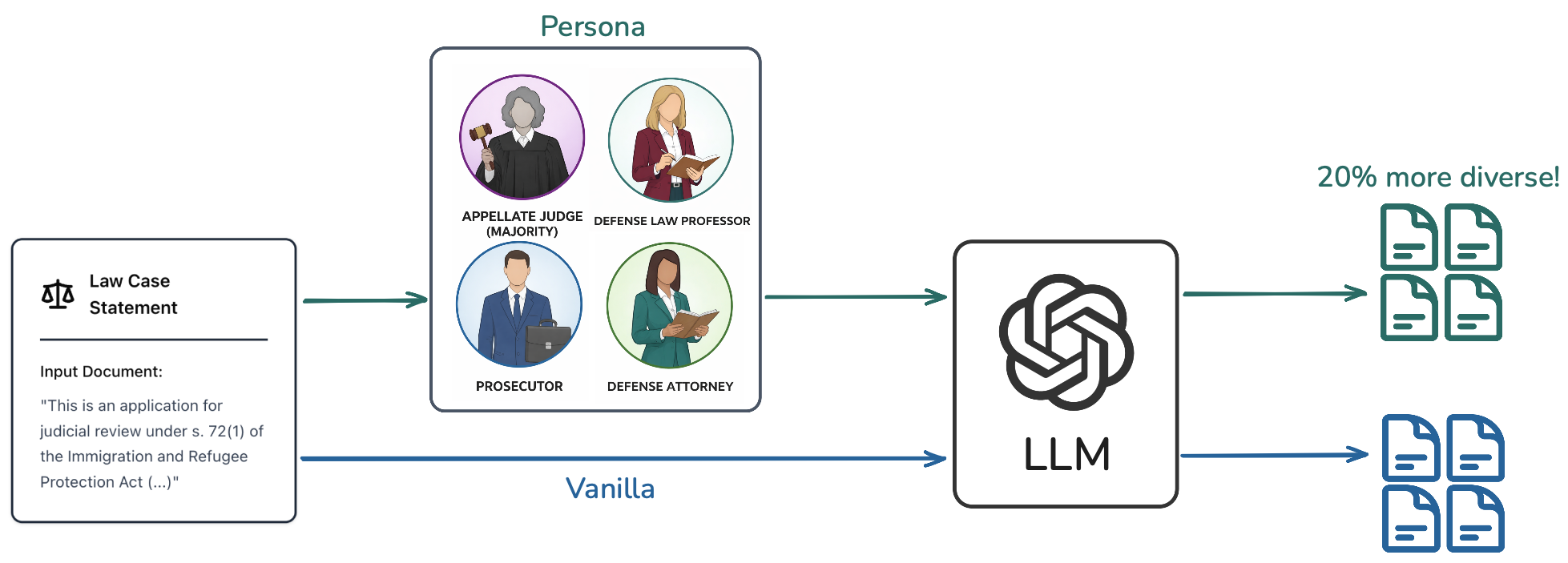}
      \caption{The DALDALL framework for persona-based data augmentation.
      Given a legal case statement, the \textbf{Persona} method (top) generates synthetic queries
      from multiple professional perspectives (attorney, prosecutor, judge, law professor),
      while the \textbf{Vanilla} method (bottom) does not incorporate persona information.
      Compared to \textbf{Vanilla}, \textbf{Persona} augmentation produces outputs with
      20\% greater lexical diversity (Table~\ref{tab:token-length-diversity}).}
      \label{fig:fig1}
\end{figure*}

Information retrieval (IR) in low-resource domains has long suffered from
data scarcity~\citep{fadaee2017data, fang2022lowresource}. The advent of large language models (LLMs) such as GPT~\citep{brown2020language}
has provided a highly convenient means of generating augmented datasets~\citep{anaby2020deep, dai2025auggpt},
leading to numerous approaches that leverage the generative capabilities
of LLMs to create AI-labeled data from limited seed examples. Among these,
few-shot prompting is widely used~\citep{bonifacio2022inpars, dai2022promptagator, jeronymo2023inparsv2, krastev2025inparsplus}, where seed data and their labels serve
as in-context examples within a prompt. However, few-shot examples bias
the model toward reusing vocabulary from the provided examples, reducing
the diversity of augmentations. This raises concerns about the quality of
augmented data in low-resource domains~\citep{rachmat2025qa}. Furthermore, while many data
augmentation methods have demonstrated the ability to generate large
volumes of synthetic samples, domain-specific approaches---particularly
for low-resource domains---remain largely underexplored~\citep{jeong2025llmbased, seo2024retrieval}. In such settings,
generating high-quality data is often more important than producing large
quantities of mediocre samples, especially in specialized domains such as
law and medicine.

In this work, we focus on the legal domain. Legal IR is challenging because of the limited availability
of high-quality annotated data~\citep{hong2020legal, kim2025gure}. To address this challenge, we employ persona-based prompting to generate
high-quality augmentations. By assigning a specific role or persona to the model---for example, simulating
a client's behavior by adopting their perspective---LLMs can exhibit more accurate and structured reasoning
compared to generic prompting~\citep{luo2024personamath}. Prior work suggests that the role assigned to an LLM is a significant factor
in shaping its outputs. Crucially, the legal domain features well-defined professional roles---such as attorneys,
prosecutors, and judges---each with distinct perspectives and linguistic patterns when formulating queries.
This makes persona-based prompting a natural fit for generating diverse synthetic queries.

As a low-resource domain, legal IR requires quality-focused approaches to augment datasets.
We decompose quality in the IR context into two dimensions: lexical diversity and semantic diversity.
Lexical diversity is important for sparse retrieval: as many practical search systems still rely on
sparse retrieval methods such as BM25. Dense retrievers benefit from exposure to diverse semantic patterns
during training and are widely used for re-ranking documents retrieved by sparse retrievers. Based on this
perspective, we propose a data augmentation framework that leverages the generative capabilities of LLMs and
introduces a persona-based augmentation strategy to induce substantial lexical diversity in legal queries (Figure \ref{fig:fig1}).
We analyze both the diversity characteristics and the downstream retrieval effectiveness of our approach.

Our contributions are as follows.
\vspace{-0.8\baselineskip}
\begin{enumerate}
      \setlength{\itemsep}{0pt}
      \setlength{\parskip}{0pt}
      \setlength{\topsep}{0pt}
      \item We introduce a persona-based augmentation strategy using legal-domain-specific professional roles,
            which induces substantial lexical and semantic diversity, as measured by Self-BLEU scores and intra-cosine similarity.
      \item We provide an empirical analysis of the conditions under which persona-based prompting yields diverse outputs and
            demonstrate its effectiveness through improved or maintained recall scores after fine-tuning multiple retrieval models.
\end{enumerate}

\section{Related Works}

\subsection{Legal Information Retrieval}

Legal IR focuses on finding relevant legal
documents---such as case law, statutes, and precedents---given a user query.
Legal IR presents unique challenges that distinguish it from general-domain
retrieval. First, legal case statements are often extremely long, frequently
exceeding the context limits of commonly used retrieval models
\citep{nguyen2025nowj}. This necessitates summarization or chunking strategies
\citep{tran2020encoded} that may lose critical information. Second, legal
language is highly specialized, featuring domain-specific terminology, complex
sentence structures, and references to statutes and precedents that require
expert understanding \citep{hong2020legal}. Third, and most critically for this
work, high-quality annotated datasets for legal IR are scarce, as creating such
datasets requires expensive legal expertise.

In response to these challenges, researchers have proposed various approaches.
Dense embedding approaches have shown promising results by encoding semantic similarity.
Other works focus on query reformulation~\citep{kim2025gure} or leverage transfer learning from
related domains~\citep{fang2022lowresource}.

\subsection{LLM-based Data Augmentation and Persona-based Prompting}

\subsubsection{LLM-Based Data Augmentation}

The emergence of large language models (LLMs) has transformed data augmentation
by enabling high-volume generation of synthetic training data. LLMs such as the
GPT family \citep{brown2020language} demonstrate remarkable capabilities to
generate synthetic data of reasonable quality in large volumes
\citep{anaby2020deep}. In the IR domain specifically, several works leverage
LLMs to generate synthetic queries for documents. InPars \citep{bonifacio2022inpars}
pioneered this approach by using few-shot prompting to generate queries, followed
by InPars-v2 \citep{jeronymo2023inparsv2} and InPars+ \citep{krastev2025inparsplus},
which refined the generation and filtering pipeline. Promptagator
\citep{dai2022promptagator} demonstrated that effective dense retrievers can be
trained from as few as eight examples. AugGPT \citep{dai2025auggpt} further
explored ChatGPT's potential for general text data augmentation.

In specialized domains such as law and medicine, generating high-quality data is often more important than
producing large volumes of mediocre samples \citep{rachmat2025qa}. This has led to more
sophisticated augmentation pipelines that incorporate retrieval-augmented generation
(RAG) \citep{seo2024retrieval} or domain-specific filtering mechanisms.

Given the importance of prompt design in LLM-based augmentation, we
now turn to a specific prompting strategy that has shown particular promise:
persona-based prompting.

\subsubsection{Persona-Based Prompting}

Prompt engineering has emerged as a critical technique for eliciting desired
behaviors from LLMs. Research has shown that even minor modifications to prompts
can lead to substantial performance improvements \citep{radford2021learning}.
This sensitivity makes prompt design a crucial factor in any LLM-based pipeline,
including data augmentation.

One influential direction in prompt engineering is persona-based prompting,
which involves assigning a specific role or identity to the LLM before it
performs a task.
This finding suggests that personas can guide LLMs toward more structured and
domain-appropriate outputs.

The persona methodology has been extended to various applications. Large-scale
persona datasets, such as Persona Hub with one billion personas \citep{ge2024scaling},
enable diverse synthetic data generation across domains. For data augmentation
specifically, recent works show that persona-driven approaches can improve both
the quality and diversity of generated text \citep{jeong2025llmbased}. The
Debate-to-Write framework \citep{hu2025debate} uses multiple personas to generate
diverse arguments, demonstrating that different perspectives yield lexically and
semantically varied outputs. PersonaMath \citep{luo2024personamath} demonstrated
that LLMs achieve higher accuracy on mathematical problems when using persona-based
data augmentation for the training dataset.

These findings are particularly relevant to legal IR. The legal domain features
well-defined professional roles---such as attorneys, prosecutors, and judges---each
with distinct perspectives, objectives, and linguistic patterns when formulating
queries. Our work addresses this gap by leveraging legal personas to induce lexical diversity in synthetic query generation.

\subsection{Datasets for Legal Information Retrieval}

Several benchmark datasets have been developed to advance research in legal
IR. In this section, we review two prominent datasets that
are particularly relevant to our work: COLIEE and CLERC.

\paragraph{COLIEE.} The Competition on Legal Information Extraction/Entailment
\cite{nguyen2025nowj} is an annually held competition that has become a
cornerstone benchmark for legal IR research. Now in its eleventh edition, COLIEE
comprises four primary tasks spanning case law and statute law. Task~1 focuses
on legal case retrieval, where systems must identify relevant precedent cases
(``noticed cases'') given a query case, which we employ for our seed data.
Based on the COLIEE 2024 statistics, the dataset contains a total of 7,350 case law
files, with 5,616 cases in the labeled training set, of which 1,278 serve as
query cases.

\paragraph{CLERC.} The Case Law Evaluation and Retrieval Corpus \cite{hou2025clerc}
is a recently introduced large-scale dataset designed to support both legal case
retrieval and retrieval-augmented generation tasks. Developed in collaboration
with legal professionals, CLERC is built upon the Caselaw Access Project (CAP),
which contains over 1.84 million U.S.\ federal case documents with more than
20.7 million citations. Unlike COLIEE, which uses case-to-case retrieval, CLERC
frames the task as citation retrieval---finding precedent cases that should be cited to support a given piece of legal analysis.

\section{Method \& Approach}

\afterpage{%
\begin{figure}[t]
  \centering
  \includegraphics[width=\columnwidth]{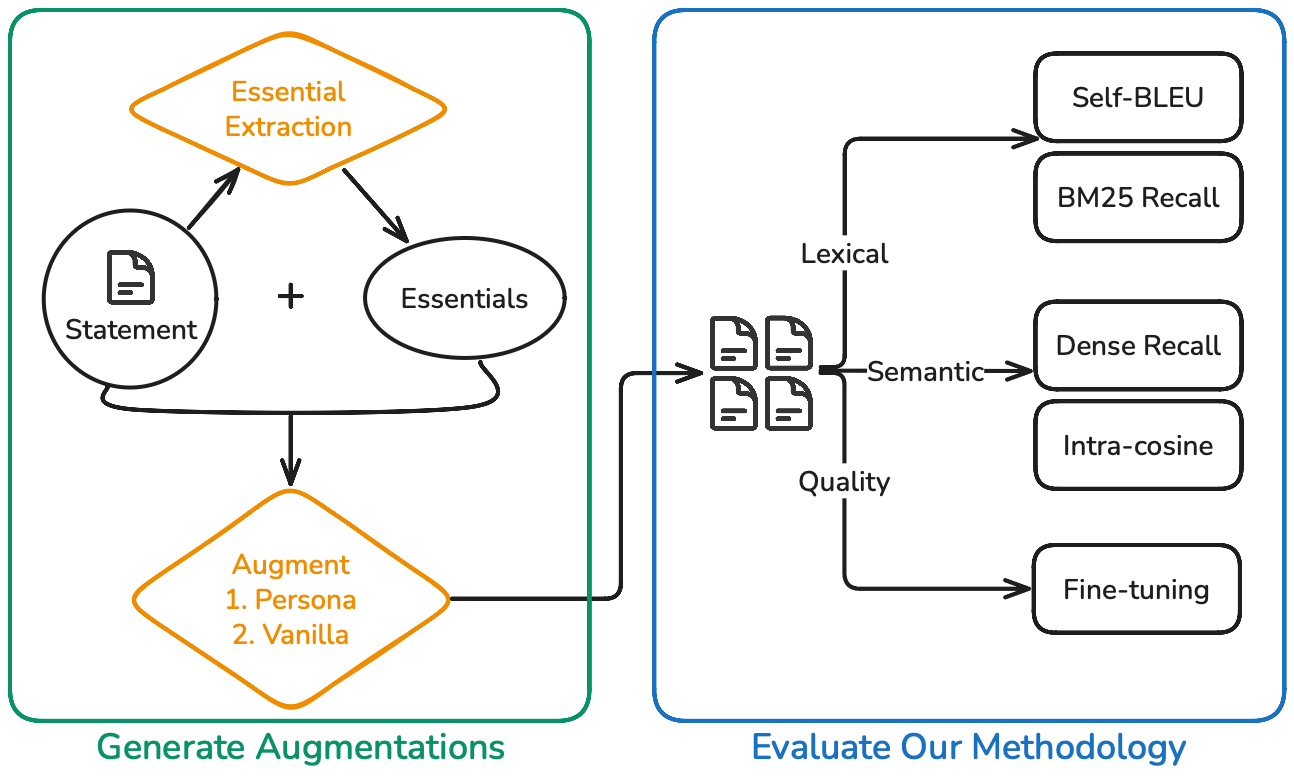}
  \caption{Overview of our experiment. \textbf{Left:} Augmentation generation pipeline,
  where the original statement and extracted essentials are combined and processed through
  either Persona or Vanilla augmentation. \textbf{Right:} Evaluation framework measuring
  lexical diversity (Self-BLEU, BM25 Recall), semantic diversity (Dense Recall, Intra-cosine),
  and augmentation quality via fine-tuning.}
  \label{fig:experiment-overview}
\end{figure}
}


\subsection{DALDALL: Persona-Based Data Augmentation}

We propose DALDALL (Data Augmentation for Lexical and Semantic Diversity in the Legal Domain by Leveraging LLM-Persona),
a data augmentation methodology tailored for legal IR. Given a query case $Q$ and its associated positive cases $D_1, D_2, \ldots$,
benchmark datasets such as CLERC and COLIEE provide labeled pairs of the form $(Q, \{D_1, D_2, \ldots\})$.

Our approach generates synthetic queries $A = f(Q, P)$ using a prompt $P$ while preserving the original positive labels.
For each query $Q$, we employ two types of prompts. The first is the \textbf{Vanilla Prompt} $P_V$,
which yields augmentation $A_V$. The second is the \textbf{Persona Prompt} $P_P$, which yields augmentation $A_P$.
The key distinction is that $P_P$ incorporates persona information absent from $P_V$. In this work, we compare
$A_P$ and $A_V$ in terms of lexical and semantic diversity as well as downstream retrieval performance.
Through this comparison, we evaluate the effectiveness of persona-based augmentation, which we hypothesize to be
more beneficial for low-resource domains than few-shot approaches that generate synthetic queries by reusing vocabulary from the examples.
For the overview of our experiment, see Figure \ref{fig:experiment-overview}.

We use five personas for comparison and fine-tuning, as our ablation study (Section~\ref{sec:ablation})
demonstrates that this subset yields optimal performance. Full descriptions of the personas and prompt templates
are provided in Appendix~\ref{app:prompt-templates}.


\subsection{Prompt Design}
\label{sec:prompts}

We provide the full prompt templates in the Appendix \ref{app:prompt-templates}. The augmentation process comprises two stages: essential extraction and query augmentation.

\paragraph{Stage 1: Essential Extraction.}
We first extract four essential components from the original query: legal issue,
legal test or standard, key precedents, and key statutes or rules.
These components correspond to the core elements of legal analysis as formalized
in the IRAC framework (Issue, Rule, Application, and Conclusion), which is the
standard methodology for legal reasoning. Recent work on rhetorical role segmentation
in legal NLP has similarly identified these elements as semantically distinct
units within legal documents~\citep{malik2022semantic, bhattacharya2019rhetorical}.
By extracting these invariant components before
augmentation, we ensure that the generated text preserves the legal meaning of
the source while allowing lexical and structural variation in non-essential
elements~\citep{shorten2021text}.

\paragraph{Stage 2: Augmentation.}
Given the original text, extracted essentials, and an augmentation prompt,
the LLM generates synthetic queries. We employ two types of prompts:
\vspace{-0.6\baselineskip}
\begin{itemize}
    \setlength{\itemsep}{0pt}
    \setlength{\parskip}{0pt}
    \setlength{\topsep}{0pt}
    \item \textit{Vanilla prompts} provide augmentation instructions without additional context (Section~\ref{app:vanilla}).
    \item \textit{Persona prompts} include instructions for the LLM to adopt a specific professional perspective (Section~\ref{app:persona-prompt}).
\end{itemize}

\paragraph{Generation Strategy.}
Each augmentation is generated independently using a single prompt per query.
We observed that generating all augmentations simultaneously with a single prompt
resulted in lower diversity compared to generating them individually.


\subsection{Evaluation Metrics}

We evaluate augmented data along two complementary dimensions: \textbf{lexical diversity}
and \textbf{semantic quality}. Additionally, we assess downstream utility through retrieval performance.

\vspace{-0.8\baselineskip}
\paragraph{Lexical Diversity.} We measure lexical diversity using two metrics:
\vspace{-0.6\baselineskip}
\begin{enumerate}
  \setlength{\itemsep}{0pt}
  \setlength{\parskip}{0pt}
  \setlength{\topsep}{0pt}
\item \textit{Self-BLEU} \cite{zhu2018texygen}. This metric measures similarity among generated samples; lower scores indicate greater diversity, which is desirable for training robust models.
  \item \textit{BM25 recall comparison}. We compare BM25 recall scores between augmented and original queries to verify that augmented queries are not restricted to a narrow vocabulary.
\end{enumerate}

\vspace{-1.0\baselineskip}
\paragraph{Semantic Diversity.} We assess whether augmented queries preserve the meaning of the originals:
\vspace{-0.6\baselineskip}
\begin{enumerate}
  \setlength{\itemsep}{0pt}
  \setlength{\parskip}{0pt}
  \setlength{\topsep}{0pt}
  \item \textit{Intra-Cosine Similarity}. We measure the cosine similarity among augmented queries, to ensure they span a diverse region of the semantic space, rather than clustering around a single representation.
  \item \textit{Retrieval score consistency}. We confirm that retrieval scores obtained with augmented queries are comparable to those of the originals, demonstrating effective preservation of semantic content.
\end{enumerate}

\vspace{-1.0\baselineskip}
\paragraph{Downstream Retrieval Performance.} We show that a retriever fine-tuned on augmented queries achieves performance equal to or better than one trained on original data alone.


\subsection{Fine-tuning}
\label{sec:finetuning}

\subsubsection{Dataset Preprocessing}

We fine-tune retrievers on two datasets: CLERC and COLIEE (see Table~\ref{tab:training-samples}
in Appendix~\ref{app:config} for the number of training samples per configuration). These datasets
differ significantly in text length, requiring different preprocessing approaches.

\paragraph{COLIEE.}
COLIEE contains full legal case statements with an average token count exceeding 6,000,
requiring passage-based segmentation. We segment both documents and queries into passages.
Segmenting the 1,000 original queries yields over 5,000 query-passage pairs, which matches
the scale of the 5,000 augmentations per method.

\paragraph{CLERC.}
Queries in CLERC average 347 tokens, well within the context window of most dense retrievers.
We use the original queries and augmentations directly without segmentation.
The training set consists of 1,000 original queries and 5,000 augmentations per method (Vanilla and Persona).

\subsubsection{Passage-Based Retrieval}
\label{sec:passage-retrieval}

To handle COLIEE's long documents, we adopt a passage-based retrieval approach.
Following \cite{nguyen2025nowj}, we consider three passage-based scoring
approaches: \textbf{FirstP}, \textbf{MaxP}, and \textbf{SumP}. We adopt MaxP,
which selects the highest similarity score among query-passage pairs $(Q, P_i)$,
where $P_i$ denotes the $i$-th passage in the document. MaxP is particularly
well-suited for legal IR because key legal reasoning is often localized within
a specific paragraph, such as the holding or reasoning section of a judgment.

In addition to segmenting documents, we also segment query cases into passages. ColBERT introduces the Late Interaction mechanism, which aggregates token-level similarities as follows:
\begin{equation}
    \text{score}(d) = \sum_{i} \max_{j \in d} \text{sim}(q_i, d_j)
\end{equation}

A related variant, the Global Max approach, computes:
\begin{equation}
    \text{score}(d) = \max_{i,j} \text{sim}(q_i, d_j)
\end{equation}

We adopt Global Max for evaluation, as preliminary experiments on the COLIEE test set using BGE-base-en-v1.5
with default configurations showed that it provides stable baseline recall performance. During evaluation on
the COLIEE dataset, to compute Recall@$k$, we calculate similarity scores between all query chunks and all
document passages, selecting the globally maximum similarity pair for each query-document comparison.

\begin{figure}[t]
  \centering
  \includegraphics[width=\linewidth]{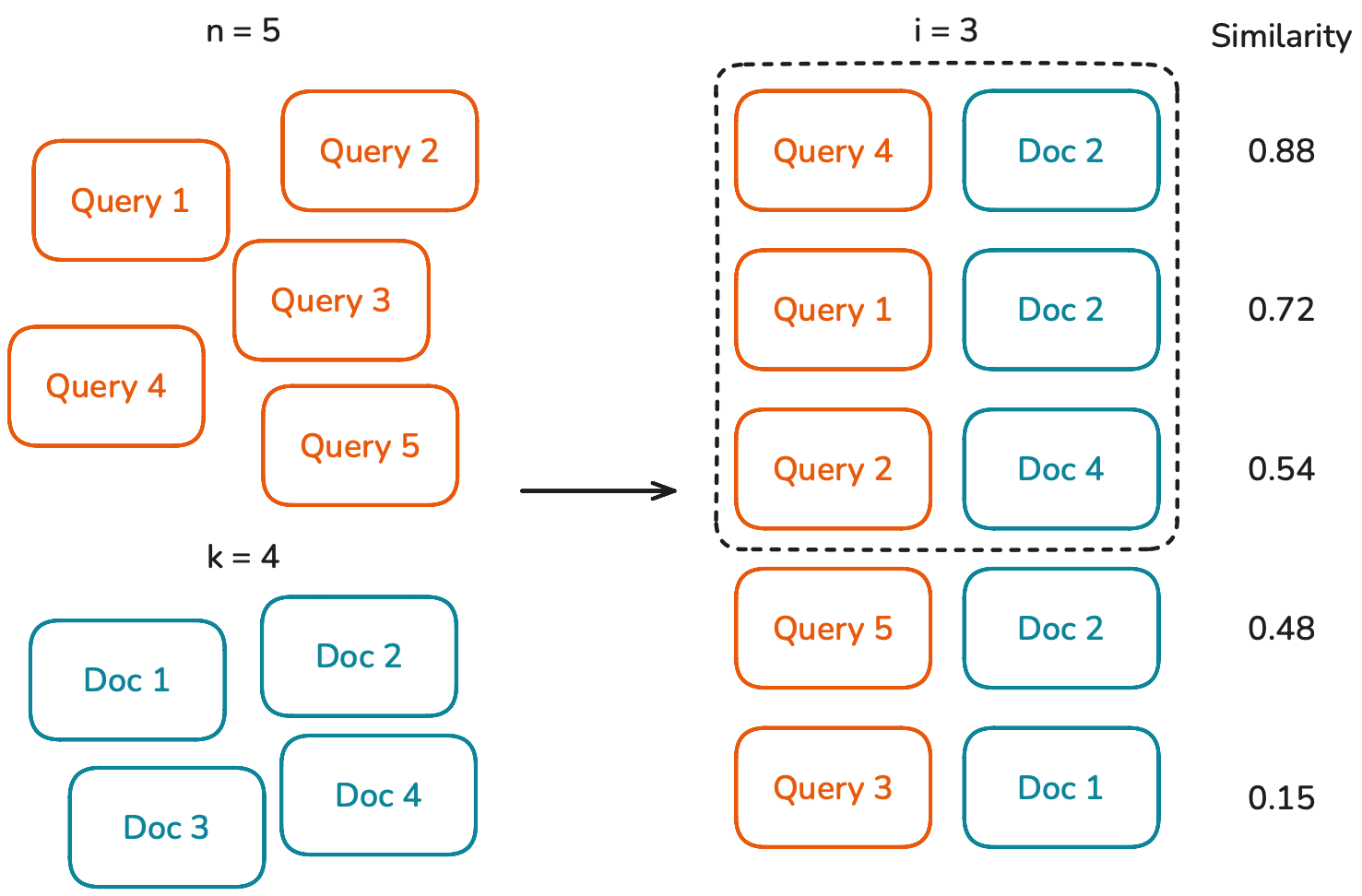}
  \caption{Triplet composition for fine-tuning on the COLIEE dataset. We chunk a query and positive documents,
  and find k-best semantic pair with base model. If $n$ is lower than $i$, then all query chunks are used for the triplet.}
  \label{fig:fig3}
\end{figure}

\subsubsection{Triplet Construction}
\label{sec:triplet}

\paragraph{COLIEE.}
Each query $Q$ is segmented into $n$ chunks: $Q_1, Q_2, \ldots, Q_n$. For each query, there are $m$ positive documents $D_1, \ldots, D_m$ and $i$ augmentations $A_1, \ldots, A_i$, where $i \in \{3, 5, 7, 10\}$ in our experiments. Since each augmentation is under 500 tokens, no further segmentation is required for the augmented queries.

Prior to fine-tuning, we construct training triplets with two objectives: \textbf{(1)} ensuring
that the number of triplets derived from original query chunks does not exceed the number of
augmentation-based triplets, and \textbf{(2)} reducing training noise by excluding low-relevance
chunk pairs. Specifically, we select the top-$i$ query-chunk--document-chunk pairs ranked by
baseline dense retriever recall score. If $n < i$, all available chunks are included without
repetition, rather than artificially padding to reach $i$. The goal of this setup is not to
demonstrate that the augmentation-based training labels outperform the original ones, but rather
to verify that the augmentations do not distort the model's learned representation---thereby
confirming that the persona-based augmentation approach contributes a meaningful, complementary training signal.
Generating the triplets for fine-tuning is illustrated in Figure \ref{fig:fig3}.

\paragraph{CLERC.}
Since CLERC queries require no segmentation, triplet construction is straightforward: each original
query or augmentation is paired directly with its positive and negative documents.

\subsubsection{Training Configurations}
\label{sec:training-config}
We evaluate five dense retrievers: DPR, LegalBERT, BGE-base-en-v1.5, multilingual-e5-base, and BGE-M3, along with BM25 as a sparse baseline. For each dense retriever, we consider six configurations:
\vspace{-0.8\baselineskip}
\begin{enumerate}
    \setlength{\itemsep}{0pt}
    \setlength{\parskip}{0pt}
    \setlength{\topsep}{0pt}
    \item \textbf{Baseline:} No fine-tuning
    \item \textbf{Original:} Original Queries
    \item \textbf{Vanilla only:} Vanilla augmentations
    \item \textbf{Vanilla mix:} Original + vanilla augmentations
    \item \textbf{Persona only:} Persona augmentations
    \item \textbf{Persona mix:} Original + persona augmentations
\end{enumerate}
\vspace{-0.8\baselineskip}

For ``mix'' configurations, we randomly sample from the combined pool of original
and augmented queries to maintain the same training set size.

\begin{table}[t]
    \centering
    \small
    \resizebox{\columnwidth}{!}{%
    \begin{tabular}{l l c c c}
        \toprule
        \textbf{Dataset} & \textbf{Approach} & \textbf{Token Len. Avg.} & \textbf{Self-BLEU} & \textbf{Intra-Cos} \\
        \midrule
        \textbf{CLERC} & Original & 347.7 & -- & -- \\
         & Persona & 124.83 & \textbf{0.368} & 0.855 \\
         & Vanilla & 86.14 & 0.435 & \textbf{0.819} \\
        \addlinespace
        \textbf{COLIEE} & Original & 6,379.8 & -- & -- \\
         & Persona & 118.9 & \textbf{0.318} & \textbf{0.778} \\
         & Vanilla & 79.2 & 0.414 & 0.799 \\
        \bottomrule
    \end{tabular}%
    }
    \caption{Lexical and semantic diversity metrics for persona-based and vanilla augmentations on CLERC and COLIEE. Each method generates 5,000 augmented queries per dataset.}
    \label{tab:token-length-diversity}
\end{table}

\section{Experiments \& Analysis}

\subsection{Various Perspectives in seed data (CLERC vs COLIEE)}

We hypothesize that each persona maintains its own standpoint on a case,
and that when there is insufficient content, the persona cannot generate diverse
perspectives. To verify this hypothesis, we conduct augmentation experiments on two
datasets: CLERC and COLIEE. And we use gpt-5-nano model for augmentation. A CLERC query contains only a small portion of
the legal case passage, focusing particularly on the citation section.
In contrast, a COLIEE query comprises the full case statement, encompassing various
perspectives and providing richer context to the LLM. We select 1,000 queries from
each dataset and generate five augmentations per query. The average token length of
CLERC queries is 338 tokens, whereas COLIEE queries average 6,380 tokens—approximately 19 times longer (Table \ref{tab:token-length-diversity}).

\paragraph{Lexical Diversity.} Contrary to our hypothesis,
persona-based augmentation consistently produces more lexically
diverse outputs than vanilla augmentation across both datasets,
as measured by Self-BLEU scores (lower indicates greater diversity) (Table \ref{tab:token-length-diversity}).
On CLERC, $P_P$ achieves a Self-BLEU of 0.368 compared to $P_V$'s 0.435,
representing a 15\% improvement in diversity. On COLIEE, this gap widens further,
with Persona at 0.318 versus $P_V$ at 0.414—a 23\% improvement.
This consistency across datasets with vastly different source text lengths
(338 vs. 6,380 tokens) suggests that the diversity gains stem primarily
from the introduction of distinct professional perspectives rather than
from characteristics of the input text. Legal personas naturally emphasize
different aspects of case documents, including lexical variation that $P_V$
prompting cannot achieve through simple paraphrasing.

\begin{table}[t]
    \centering
    \setlength{\tabcolsep}{4pt}
    \renewcommand{\arraystretch}{1.1}
    \resizebox{\columnwidth}{!}{%
    \begin{tabular}{c cc cc}
        \toprule
        \textbf{Section} &
        \multicolumn{2}{c}{\textbf{Persona}} &
        \multicolumn{2}{c}{\textbf{Vanilla}} \\
        \cmidrule(lr){2-3} \cmidrule(lr){4-5}
         & \textbf{Self-BLEU} & \textbf{Intra-Cos}
         & \textbf{Self-BLEU} & \textbf{Intra-Cos} \\
        \midrule
        1 & 0.321 & 0.785 & 0.463 & 0.818 \\
        2 & 0.313 & 0.778 & 0.429 & 0.803 \\
        3 & 0.300 & 0.774 & 0.421 & 0.795 \\
        4 & 0.300 & 0.770 & 0.410 & 0.782 \\
        5 & 0.299 & 0.771 & 0.395 & 0.781 \\
        6 & 0.290 & 0.769 & 0.391 & 0.786 \\
        7 & 0.282 & 0.765 & 0.379 & 0.779 \\
        \bottomrule
    \end{tabular}
    }
    \caption{Diversity metrics by document length on COLIEE. Documents are divided into seven sections of 1,000, ordered by token count (Section 1 = shortest). Lower Self-BLEU and Intra-Cosine scores indicate greater diversity.}
    \label{tab:section-diversity}
\end{table}

\paragraph{Semantic Diversity.} Intra-Cosine Similarity measures semantic variation
among augmented queries, where lower scores indicate greater semantic diversity (Table \ref{tab:section-diversity}).
On CLERC, $P_P$ achieves a higher Intra-Cosine Similarity (0.855) than $P_V$ (0.819),
indicating that persona-based augmentations are semantically more similar to one another
despite being lexically more diverse. On COLIEE, however, this pattern reverses:
$P_P$ produces a lower Intra-Cosine Similarity (0.778) than $P_V$ (0.799),
demonstrating greater semantic diversity. We attribute this difference to the
length of the source texts. Shorter CLERC queries provide limited semantic content
for personas to differentiate, resulting in augmentations that vary in wording but
converge in meaning. In contrast, longer COLIEE documents offer sufficient material
for each persona to emphasize genuinely distinct aspects of the case, enabling both
lexical and semantic diversification. This finding partially supports our hypothesis:
input text richness influences semantic diversity, even though lexical diversity
improvements remain consistent regardless of source length.

\subsection{Token Distribution}

\begin{table}[t]
    \centering
    \small
    \begin{tabular}{l c c c c c}
        \toprule
        \textbf{Approach} & \textbf{R@1} & \textbf{R@5} & \textbf{R@10} & \textbf{R@20} & \textbf{R@50} \\
        \midrule
        \textbf{Original} & 0.0540 & 0.1784 & 0.2572 & 0.3348 & 0.4481 \\
        \textbf{Vanilla} & 0.0438 & 0.1406 & 0.2092 & 0.2894 & 0.4035 \\
        \textbf{Persona} & 0.0410 & 0.1360 & 0.1973 & 0.2690 & 0.3729 \\
        \bottomrule
    \end{tabular}
    \caption{BM25 recall on COLIEE. 1,000 original queries are sampled, with 5 augmentations generated per query (5,000 total per method).}
    \label{tab:bm25-retrieval-coliee}
\end{table}

\begin{table}[t]
    \centering
    \small
    \begin{tabular}{l c c c c c}
        \toprule
        \textbf{Approach} & \textbf{R@1} & \textbf{R@5} & \textbf{R@10} & \textbf{R@20} & \textbf{R@50} \\
        \midrule
        \textbf{Original} & 0.0271 & 0.0887 & 0.1280 & 0.1827 & 0.2665 \\
        \textbf{Vanilla} & 0.0250 & 0.0708 & 0.1022 & 0.1438 & 0.2149 \\
        \textbf{Persona} & 0.0208 & 0.0604 & 0.0884 & 0.1241 & 0.1890 \\
        \bottomrule
    \end{tabular}
    \caption{Retrieval performance of BGE-base-en-v1.5 (non-fine-tuned) on COLIEE. Dataset sizes are identical to Table 4.}
    \label{tab:bge-base-non-ft}
\end{table}

We further investigate the initial hypothesis that longer input texts can
generate more diverse augmentations. Notably, a pattern emerges in the results
of CLERC versus COLIEE: both Self-BLEU and Intra-Cosine Similarity scores for both
$P_V$ and $P_P$ on COLIEE are better than on CLERC.

The COLIEE dataset exhibits a wide token distribution across 7,350 texts.
The 1,000th document contains 2,089 tokens, while the 7,000th document contains
15,798 tokens. We divide the dataset into seven sections, each containing 1,000 texts.
Each section index represents a ranking range: Section 1 corresponds to documents
ranked 1-1,000; Section 2 corresponds to ranks 1,001-2,000; and so forth.

The more tokens the input texts contain, the lower the Self-BLEU scores,
regardless of approach (Table \ref{tab:section-diversity}). Lexical diversity steadily improves when the
original text has ample material for augmentation. Additionally, all
sections demonstrate that the Persona approach can generate more diverse
augmentations by producing more unique tokens and exploiting a wider
variety of words.

\subsection{Augmentation Quality}

Our methodology transforms an original query into short augmentations that can be utilized by compact retrieval models such as DPR and BERT-based encoders. This is advantageous for IR because compact models remain widely adopted in this field due to their computational efficiency. However, we must demonstrate that the generated queries are sufficiently representative of the original query. This raises the following research questions:

1. Do the augmentations retain an appropriate degree of lexical overlap with the original query—neither too narrow (under-retrieving relevant passages) nor too broad (over-retrieving irrelevant ones)?
2. Do the augmentations preserve semantic proximity to the original query's representation? Do their distributional shifts effectively enhance retrieval performance?

We conduct experiments to validate the quality of our augmentations on COLIEE with respect to these questions.

\vspace{-0.8\baselineskip}
\begin{enumerate}
    \setlength{\itemsep}{0pt}
    \setlength{\parskip}{0pt}
    \setlength{\topsep}{0pt}
    \item \textbf{BM25 retrieval scores:} We evaluate whether the augmentations preserve sufficient lexical overlap with positive passages without excessive keyword dilution.
    \item \textbf{BGE-base retrieval scores:} We assess whether the augmentations retain semantic proximity to the original query in the embedding space.
    \item \textbf{Fine-tuning and evaluation:} We verify that the semantic shifts induced by the augmentations lead to meaningful improvements in retrieval performance.
\end{enumerate}

\begin{table}[t]
    \centering
    \caption{Semantic similarity metrics for COLIEE augmentations.}
    \label{tab:semantic-similarity}
    \small
\begin{tabular}{l l c c}
        \toprule
        \textbf{Dataset} & \textbf{Approach} & \textbf{Avg} & \textbf{Mean} \\
        \midrule
        \textbf{COLIEE} & Persona & 0.697 & 0.843 \\
         & Vanilla & 0.682 & 0.840 \\
        \bottomrule
    \end{tabular}
\end{table}

Lexical overlap decreases for both augmentation methods,
with a more pronounced reduction in the Persona approach due
to its generation of more lexically diverse terms (Table \ref{tab:semantic-similarity}). Nevertheless,
the difference from the original queries remains within approximately
20\%, indicating that the augmentations preserve key lexical features
adequately. This suggests that the augmented data can serve as a viable
legal IR dataset with (augmented query, positive passage) pairs that
are sufficiently challenging for model training.

We analyze two aspects: BGE-base-en-v1.5 recall scores and semantic
similarity metrics. Although the Persona and Vanilla methods exhibit
similar semantic proximity to the original queries, the dense retrieval
results reveal that their augmentations occupy distinct regions in the
embedding space, suggesting that semantic similarity alone may not
predict retrieval performance (Table \ref{tab:semantic-similarity}). To investigate whether the semantic
space of Persona-augmented queries aligns better with the retriever’s
learned representations, we further examine fine-tuning performance.

DPR and BGE-base-en-v1.5 achieve better performance (Table~\ref{tab:finetune-retrievers} in Appendix~\ref{app:retrieval-results}) with Persona augmentations, even though the evaluation setup favors the Original method (i.e.,
chunking original case sentences and using them directly for retrieval). In contrast, LegalBERT,
BGE-M3, and E5-base with Persona do not outperform the other methods. We hypothesize that LegalBERT,
having been pre-trained on legal texts such as legislation, court cases, and contracts, is less
compatible with the more informal and naturally phrased augmentations, which may interfere with
its domain-specific representations. Additionally, BGE-M3 has a larger context window, which allows
it to incorporate more content from the original cases; this may explain its higher recall scores
compared to those of the Persona method. Finally, E5-base is a general-purpose model that may not
be as effective for the legal domain as the specialized models, which could explain why there are
no significant differences among the methods.

Based on the above analysis, we conclude that the Persona method is not always the best option
for fine-tuning and retrieval performance. However, the downstream retrieval results demonstrate
that the Persona method does not cause distribution shift in the models, except for LegalBERT,
while successfully generating more diverse augmentations. We anticipate that our approach can be
beneficial when combined with supplementary methodologies for low-resource domains. Full retrieval results are provided in Table~\ref{tab:finetune-retrievers} in Appendix~\ref{app:retrieval-results}.
\section{Ablation}
\label{sec:ablation}

\subsection{How many personas are needed?}

\begin{figure}[t]
    \centering
    \includegraphics[width=\linewidth]{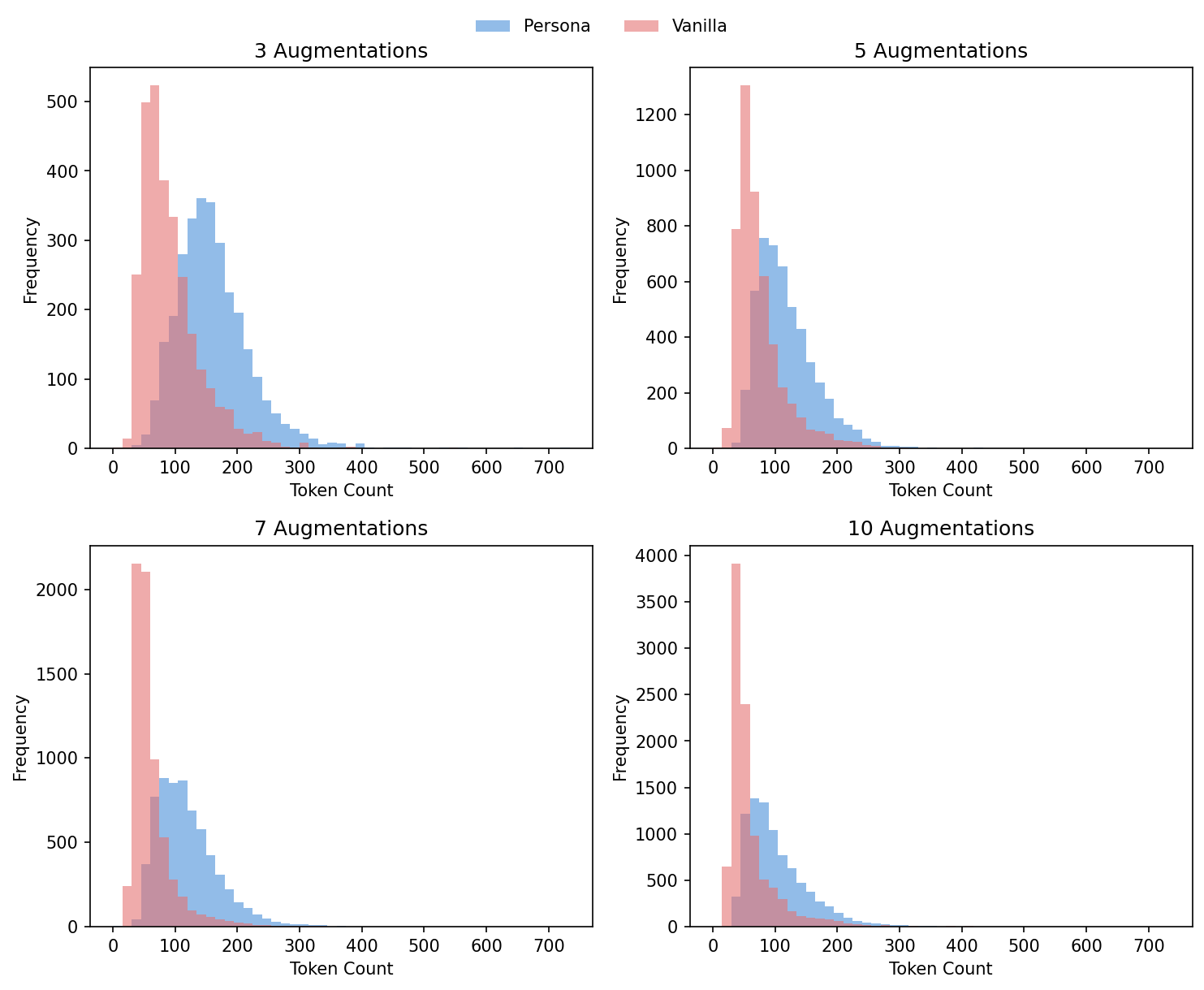}
    \caption{Distribution of augmented query lengths on COLIEE for 3, 5, 7, and 10 personas. Persona-based prompting yields a broader token distribution than vanilla prompting regardless of persona count.}
    \label{fig:fig4}
\end{figure}

We conduct an experiment to determine the optimal number of personas using the COLIEE dataset.
We evaluate four persona set sizes (3, 5, 7, and 10), where each smaller set is a nested subset
of the full 10-persona set. The composition of each set is detailed in the Appendix \ref{app:persona-comb}.

\begin{figure}[t]
    \centering
    \includegraphics[width=\linewidth]{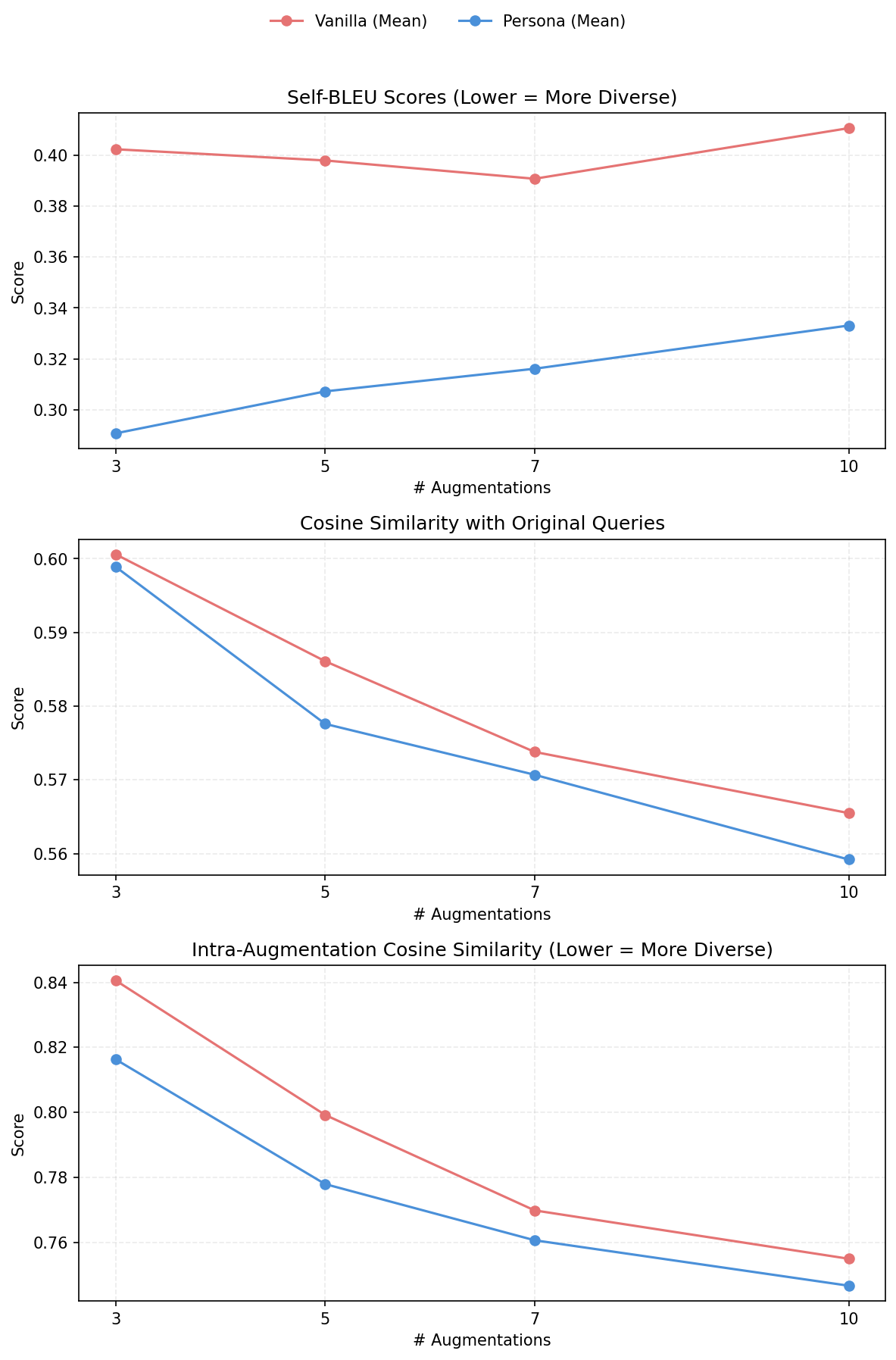}
    \caption{Effect of persona count on augmentation diversity (COLIEE). Top: Self-BLEU scores (lower = more lexically diverse). Middle: cosine similarity with original queries. Bottom: intra-augmentation cosine similarity (lower = more semantically diverse). Increasing persona count reduces lexical diversity but improves semantic diversity. Five personas offer a balanced trade-off.}
    \label{fig:fig5}
\end{figure}

Several clear trends emerge as the number of personas varies (Figure \ref{fig:fig5}). With more personas,
Self-BLEU scores rise (indicating reduced lexical diversity), while both cosine
similarity with original queries and intra-augmentation cosine similarity decrease
(indicating greater semantic diversity). Interestingly, the 3-persona configuration
exhibits a trade-off: it achieves the lowest Self-BLEU score and the largest Vanilla-Persona gap,
yet its intra-augmentation cosine similarity remains relatively high. This suggests
that fewer personas produce lexically varied augmentations that nonetheless occupy a
more constrained semantic region.

The results do not reveal a single optimal number of personas.
We therefore select five personas as a practical compromise based on the following reasoning:
\vspace{-1.0\baselineskip}
\begin{itemize}
    \setlength{\itemsep}{0pt}
    \setlength{\parskip}{0pt}
    \setlength{\topsep}{0pt}
    \item Substantial semantic diversity gain from 3 to 5 personas
    \item Sufficient Self-BLEU gap between Vanilla-Persona methods
    \item Using 5 personas is computationally cheaper than 7 or 10 while achieving competitive fine-tuning performance.
\end{itemize}
\section{Limitations \& Future Work}
Despite the growing availability of legal datasets in various languages, this
work focuses exclusively on English datasets. Future research should investigate
whether our methodology generalizes to other languages. Additionally, as we are
not legal experts and could not afford professional legal annotators, we cannot
verify that our augmentations are legally sound and reliable; however, retrieval
performance remains competitive, suggesting that the augmentations are at least
functionally useful. To mitigate this concern, we introduced the concept of
"Essentials" to guide LLM-based augmentation and prevent alteration of key legal
facts in the arguments. Furthermore, we did not explore how different
combinations of personas with a fixed set size affect the metrics. Our current
findings suggest that the number of personas influences lexical and semantic
diversity; however, future work should examine which specific personas are most
effective in the legal domain.

\section{Conclusion}

In this work, we presented DALDALL, a data augmentation methodology that enhances
lexical and semantic diversity in the legal IR domain by
leveraging the generative capabilities of LLMs. We demonstrated that LLM-based
personas significantly affect augmentation diversity, with particular
effectiveness in the legal domain. Additionally, we evaluated the effectiveness
of our augmentation approach by fine-tuning retrievers on the augmented data. We
believe this work provides a practical guide for generating augmentations in
low-resource domains.

\bibliography{daldall}

@inproceedings{anaby2020deep,
  author    = {Anaby-Tavor, Ateret and Carmeli, Boaz and Goldbraich, Esther and 
               Kantor, Amir and Kour, George and Shlomov, Segev and 
               Stern, Aviv and Zwerdling, Naama},
  title     = {Do Not Have Enough Data? {Deep} Learning to the Rescue!},
  booktitle = {Proceedings of the AAAI Conference on Artificial Intelligence},
  volume    = {34},
  pages     = {7383--7390},
  year      = {2020}
}

@article{bonifacio2022inpars,
  author  = {Bonifacio, Luiz and Abonizio, Hugo and Fadaee, Marzieh and 
             Nogueira, Rodrigo},
  title   = {{InPars}: Data Augmentation for Information Retrieval Using 
             Large Language Models},
  journal = {arXiv preprint arXiv:2202.05144},
  year    = {2022}
}

@inproceedings{brown2020language,
  author    = {Brown, Tom and Mann, Benjamin and Ryder, Nick and Subbiah, Melanie and 
               Kaplan, Jared D. and Dhariwal, Prafulla and Neelakantan, Arvind and 
               Shyam, Pranav and Sastry, Girish and Askell, Amanda and 
               Agarwal, Sandhini and Herbert-Voss, Ariel and Krueger, Gretchen and 
               Henighan, Tom and Child, Rewon and Ramesh, Aditya and 
               Ziegler, Daniel M. and Wu, Jeffrey and Winter, Clemens and 
               Hesse, Christopher and Chen, Mark and Sigler, Eric and 
               Litwin, Mateusz and Gray, Scott and Chess, Benjamin and 
               Clark, Jack and Berner, Christopher and McCandlish, Sam and 
               Radford, Alec and Sutskever, Ilya and Amodei, Dario},
  title     = {Language Models are Few-Shot Learners},
  booktitle = {Advances in Neural Information Processing Systems},
  volume    = {33},
  pages     = {1877--1901},
  year      = {2020}
}

@article{dai2025auggpt,
  author  = {Dai, Haixing and Liu, Zhengliang and Liao, Wenxiong and 
             Huang, Xiaoke and Cao, Yihan and Wu, Zihao and Zhao, Lin and 
             Xu, Shaochen and Liu, Wei and Liu, Ninghao and Li, Sheng and 
             Zhu, Dajiang and Cai, Hongmin and Sun, Lichao and Li, Quanzheng and 
             Shen, Dinggang and Liu, Tianming and Li, Xiang},
  title   = {{AugGPT}: Leveraging {ChatGPT} for Text Data Augmentation},
  journal = {IEEE Transactions on Big Data},
  year    = {2025}
}

@article{dai2022promptagator,
  author  = {Dai, Zhuyun and Zhao, Vincent Y. and Ma, Ji and Luan, Yi and 
             Ni, Jianmo and Lu, Jing and Bakalov, Anton and Guu, Kelvin and 
             Hall, Keith B. and Chang, Ming-Wei},
  title   = {Promptagator: Few-shot Dense Retrieval from 8 Examples},
  journal = {arXiv preprint arXiv:2209.11755},
  year    = {2022}
}

@article{fadaee2017data,
  author  = {Fadaee, Marzieh and Bisazza, Arianna and Monz, Christof},
  title   = {Data Augmentation for Low-Resource Neural Machine Translation},
  journal = {arXiv preprint arXiv:1705.00440},
  year    = {2017}
}

@inproceedings{fang2022lowresource,
  author    = {Fang, Jie and Li, Xiang and Liu, Yang},
  title     = {Low-Resource Similar Case Matching in Legal Domain},
  booktitle = {Artificial Neural Networks and Machine Learning -- ICANN 2022},
  editor    = {Pimenidis, Elias and Angelov, Plamen and Jayne, Chrisina and 
               Papaleonidas, Antonios and Aydin, Mehmet},
  series    = {Lecture Notes in Computer Science},
  volume    = {13530},
  pages     = {571--582},
  publisher = {Springer},
  address   = {Cham},
  year      = {2022},
  doi       = {10.1007/978-3-031-15931-2_47}
}

@article{ge2024scaling,
  author  = {Ge, Tao and Chan, Xin and Wang, Xiaoyang and Yu, Dian and 
             Mi, Haitao and Yu, Dong},
  title   = {Scaling Synthetic Data Creation with 1,000,000,000 Personas},
  journal = {arXiv preprint arXiv:2406.20094},
  year    = {2024}
}

@inproceedings{hong2020legal,
  author    = {Hong, Zhen and Zhou, Qinghua and Zhang, Rongcheng and 
               Li, Weiping and Mo, Tong},
  title     = {Legal Feature Enhanced Semantic Matching Network for 
               Similar Case Matching},
  booktitle = {2020 International Joint Conference on Neural Networks (IJCNN)},
  pages     = {1--8},
  address   = {Glasgow, UK},
  month     = jul,
  year      = {2020},
  publisher = {IEEE},
  doi       = {10.1109/IJCNN48605.2020.9207528}
}

@inproceedings{hou2025clerc,
  author    = {Hou, Aaron Berkowitz and Weller, Orion and Qin, Guanghui and 
               Yang, Eugene and Lawrie, Dawn and Holzenberger, Nils and 
               Zotkina, Elena and Van Durme, Benjamin},
  title     = {{CLERC}: A Dataset for Legal Case Retrieval and 
               Retrieval-Augmented Analysis Generation},
  booktitle = {Findings of the Association for Computational Linguistics: NAACL 2025},
  pages     = {7898--7913},
  month     = apr,
  year      = {2025}
}

@inproceedings{hu2025debate,
  author    = {Hu, Zhe and Chan, Hou Pong and Li, Jing and Yin, Yu},
  title     = {Debate-to-Write: A Persona-Driven Multi-Agent Framework for 
               Diverse Argument Generation},
  booktitle = {Proceedings of the 31st International Conference on 
               Computational Linguistics},
  pages     = {4689--4703},
  month     = jan,
  year      = {2025}
}

@article{jeong2025llmbased,
  author  = {Jeong, Hee Seung and Ko, Hyung Kyu and Park, So Yeon and 
             Kim, Tae Hyun},
  title   = {{LLM}-Based Persona-Driven Text Data Augmentation},
  journal = {IEEE Access},
  year    = {2025}
}

@article{jeronymo2023inparsv2,
  author  = {Jeronymo, Vitor and Bonifacio, Luiz and Abonizio, Hugo and 
             Fadaee, Marzieh and Lotufo, Roberto and Zavrel, Jakub and 
             Nogueira, Rodrigo},
  title   = {{InPars-v2}: Large Language Models as Efficient Dataset Generators 
             for Information Retrieval},
  journal = {arXiv preprint arXiv:2301.01820},
  year    = {2023}
}

@inproceedings{kim2025gure,
  author    = {Kim, Donghyun and Kang, Donghyeon and Kim, Jihyuk and 
               Ryu, Seungwoo and Lee, Gary},
  title     = {{GuRE}: Generative Query Rewriter for Legal Passage Retrieval},
  booktitle = {Proceedings of the Natural Legal Language Processing Workshop 2025},
  pages     = {424--438},
  address   = {Suzhou, China},
  month     = nov,
  year      = {2025},
  publisher = {Association for Computational Linguistics}
}

@article{krastev2025inparsplus,
  author  = {Krastev, Milen and Hamar, Michal and Toapanta, David and 
             Brouwers, Jules and Lei, Yuanhao},
  title   = {{InPars+}: Supercharging Synthetic Data Generation for 
             Information Retrieval Systems},
  journal = {arXiv preprint arXiv:2508.13930},
  year    = {2025}
}

@article{luo2024personamath,
  author  = {Luo, Jing and Chen, Longxu and Luo, Run and Zhu, Liang and 
             Ao, Chang and Li, Jiaming and Ren, Zhaofeng and Yang, Min},
  title   = {{PersonaMath}: Boosting Mathematical Reasoning via 
             Persona-Driven Data Augmentation},
  journal = {arXiv preprint arXiv:2410.01504},
  year    = {2024}
}

@inproceedings{nguyen2025nowj,
  author    = {Nguyen, Ha-Thanh and Nguyen, Thien-Minh and Le, Xuan-Bach and 
               Le, Thi-Kieu and Nguyen, Khanh-Hung and Nguyen, Hoang-Trung and 
               Vuong, Thi-Hai-Yen and Nguyen, Le-Minh},
  title     = {{NOWJ@COLIEE} 2025: A Multi-Stage Framework Integrating 
               Embedding Models and Large Language Models for Legal 
               Retrieval and Entailment},
  booktitle = {Proceedings of the Twentieth International Conference on 
               Artificial Intelligence and Law (ICAIL '25)},
  pages     = {506--515},
  year      = {2025},
  doi       = {10.1145/3769126.3785016}
}

@inproceedings{rachmat2025qa,
  author    = {Rachmat, Brillian Kresna and Gerald, Timothy and 
               Slb, Zulfa Zahira and Grouin, Cyril},
  title     = {{QA} Analysis in Medical and Legal Domains: A Survey of 
               Data Augmentation in Low-Resource Settings},
  booktitle = {Proceedings of the 63rd Annual Meeting of the Association 
               for Computational Linguistics (Volume 4: Student Research Workshop)},
  pages     = {1132--1144},
  month     = jul,
  year      = {2025}
}

@inproceedings{radford2021learning,
  author    = {Radford, Alec and Kim, Jong Wook and Hallacy, Chris and 
               Ramesh, Aditya and Goh, Gabriel and Agarwal, Sandhini and 
               Sastry, Girish and Askell, Amanda and Mishkin, Pamela and 
               Clark, Jack and Krueger, Gretchen and Sutskever, Ilya},
  title     = {Learning Transferable Visual Models from Natural Language Supervision},
  booktitle = {International Conference on Machine Learning},
  pages     = {8748--8763},
  publisher = {PMLR},
  month     = jul,
  year      = {2021}
}

@article{seo2024retrieval,
  author  = {Seo, Minjoon and Baek, Jinhyuk and Thorne, James and 
             Hwang, Sung Ju},
  title   = {Retrieval-Augmented Data Augmentation for Low-Resource Domain Tasks},
  journal = {arXiv preprint arXiv:2402.13482},
  year    = {2024}
}

@article{tran2020encoded,
  author  = {Tran, Vu and Le Nguyen, Minh and Tojo, Satoshi and Satoh, Ken},
  title   = {Encoded Summarization: Summarizing Documents into Continuous 
             Vector Space for Legal Case Retrieval},
  journal = {Artificial Intelligence and Law},
  volume  = {28},
  pages   = {441--467},
  year    = {2020},
  doi     = {10.1007/s10506-020-09262-4}
}

@inproceedings{zhu2018texygen,
  author    = {Zhu, Yaoming and Lu, Sidi and Zheng, Lei and Guo, Jiaxian and
               Zhang, Weinan and Wang, Jun and Yu, Yong},
  title     = {Texygen: A Benchmarking Platform for Text Generation Models},
  booktitle = {The 41st International ACM SIGIR Conference on Research \&
               Development in Information Retrieval},
  pages     = {1097--1100},
  month     = jun,
  year      = {2018}
}

@inproceedings{bhattacharya2019rhetorical,
  author    = {Bhattacharya, Paheli and Paul, Shounak and Ghosh, Kripabandhu and
               Ghosh, Saptarshi and Wyner, Adam},
  title     = {Identification of Rhetorical Roles of Sentences in
               {Indian} Legal Judgments},
  booktitle = {Legal Knowledge and Information Systems: JURIX 2019},
  editor    = {Araszkiewicz, Micha{\l} and Rodr{\'i}guez-Doncel, V{\'i}ctor},
  series    = {Frontiers in Artificial Intelligence and Applications},
  volume    = {322},
  pages     = {3--12},
  publisher = {IOS Press},
  year      = {2019},
  doi       = {10.3233/FAIA190297}
}

@inproceedings{malik2022semantic,
  author    = {Malik, Vijit and Sanjay, Rishabh and Guha, Shouvik Kumar and
               Hazarika, Angshuman and Nigam, Shubham Kumar and
               Bhattacharya, Arnab and Modi, Ashutosh},
  title     = {Semantic Segmentation of Legal Documents via Rhetorical Roles},
  booktitle = {Proceedings of the Natural Legal Language Processing Workshop 2022},
  pages     = {153--171},
  month     = dec,
  year      = {2022},
  address   = {Abu Dhabi, UAE},
  publisher = {Association for Computational Linguistics},
  url       = {https://aclanthology.org/2022.nllp-1.13}
}

@article{shorten2021text,
  author    = {Shorten, Connor and Khoshgoftaar, Taghi M. and Furht, Borko},
  title     = {Text Data Augmentation for Deep Learning},
  journal   = {Journal of Big Data},
  volume    = {8},
  number    = {1},
  pages     = {101},
  year      = {2021},
  publisher = {Springer},
  doi       = {10.1186/s40537-021-00492-0}
}
\bibliographystyle{icml2026}

\clearpage
\appendix
\section{Prompt Templates}\label{app:prompt-templates}

This appendix provides the complete prompt templates used in our data augmentation pipeline. Section~\ref{app:essentials} presents the prompt for extracting legal essentials, Section~\ref{app:vanilla} shows the base augmentation template shared by both vanilla and persona prompts, and Section~\ref{app:persona-prompt} details the persona-specific extensions. Section~\ref{app:persona-desc} describes each persona, and Section~\ref{app:persona-comb} specifies the persona combinations used in our experiments.

\subsection{Extract Essentials Prompt}\label{app:essentials}

The following prompt (Figure~\ref{fig:essentials-prompt}) instructs the model to extract the invariant semantic core from legal text, which serves as the foundation for semantically consistent augmentation.

\begin{figure*}[ht]
\begin{lstlisting}[style=prompt]
You are a legal-analysis model. Your task is to extract the invariant semantic core of a legal text. The invariant core is the minimal set of legally essential propositions that must remain unchanged across all persona-based rewrites.

STRICT CONSTRAINTS:
- Use only information explicitly present or unambiguously implied by the text.
- Do not infer motives, procedural posture, missing facts, or unstated legal theories.
- Do not add legal doctrine beyond what the text names or quotes.
- Be maximally concise; remove narrative, rhetoric, or stylistic detail.
- When uncertain, use neutral phrasing ("the court indicates...", "the text states...") rather than assumptions.
- Exclude material not necessary for legal invariance, such as:
  - dicta not tied to the holding
  - background facts irrelevant to the legal issue
  - commentary or speculation
  - redundant procedural descriptions
- Before outputting JSON, internally verify that every element is text-supported and legally essential.

Essentials to extract:

These are the only fields that matter for semantic invariance:
- legal_issue: the central legal question explicitly addressed.
- legal_test_or_standard: any doctrinal test or rule stated or quoted.
- key_precedents: cited cases only (no summaries unless text gives one).
- key_statutes_or_rules: cited statutes or rules only.

Below are examples.

(few-shot examples omitted for brevity)
\end{lstlisting}
\caption{Extract Essentials prompt template.}
\label{fig:essentials-prompt}
\end{figure*}

\subsection{Augmentation Prompt (Base Template)}\label{app:vanilla}

Both vanilla and persona-based augmentation use the base prompt template shown in Figure~\ref{fig:base-prompt}. The persona prompt extends this template with additional persona-conditioning instructions (Section~\ref{app:persona-prompt}).

\begin{figure*}[ht]
\begin{lstlisting}[style=prompt]
You are an expert legal reasoning model generating COUNTERFACTUAL TEXT REWRITES for a legal information retrieval dataset.

Your task is to produce {augmentation_count} COUNTERFACTUAL TEXT REWRITES:
- Be an alternative way of asking about the *same* underlying legal situation.
- Remain semantically equivalent with respect to Essentials (E).
- Differ substantially from the Original Text in wording, syntax, and rhetorical framing.
- Differ from the other rewritten texts to promote lexical and structural diversity.

Essentials (E) are the invariant semantic core. They MUST remain unchanged.

### Diversity Requirements Across All {augmentation_count} Rewrites

Across the {augmentation_count} augmented texts, you MUST:
- Use at least three different sentence structures:
  - e.g., a single complex sentence; two short sentences; a "Under what circumstances..." style question.
- NOT start more than one augmented text with the same first three words.
- Re-order clauses or issues differently across rewrites (some standard-first, some fact-first).

Lexical Diversity Rules:
- Avoid reusing long spans (more than 5 consecutive words) from the Original Text, except for legally indispensable terms (case names, statutes, doctrine labels).
- For each augmented text, use at least 3 paraphrases or synonyms for non-technical phrases in the Original Text (e.g., "affect the outcome" -> "alter the result").

### Strict Requirements

You MUST:
- Preserve all Essentials (E) exactly; do NOT change their meaning.
- Ensure each rewritten text would retrieve the same case/passage as the original.
- Treat this as COUNTERFACTUAL DATA GENERATION:
  - Maximize lexical and structural difference from the Original Text.
  - Change word choice, sentence structure, and rhetorical focus.
  - Avoid copying phrases from the Original Text unless they are legally indispensable terms.
- Maintain legal correctness; do NOT invent new facts, issues, or rules.

You MAY:
- Re-order clauses or issues.
- Foreground or background different parts of Essentials (E).
- Compress or paraphrase non-essential details, so long as you do not contradict Essentials (E).

### NOW PERFORM THE TASK

Essentials (E): {essentials}
Original Text: {text}

Generate {augmentation_count} COUNTERFACTUAL TEXT REWRITES that preserve Essentials (E) exactly, use substantially different phrasing, and also differ meaningfully from one another.
\end{lstlisting}
\caption{Base augmentation prompt template used by both vanilla and persona-based methods.}
\label{fig:base-prompt}
\end{figure*}

\subsection{Persona-Specific Extensions}\label{app:persona-prompt}

For persona-based augmentation, we extend the base template with the modifications shown in Figure~\ref{fig:persona-extensions}. The system instruction is updated to indicate persona conditioning, a Persona Rules section is inserted after the task description, and the closing instruction is modified to emphasize persona-based diversity.

\begin{figure*}[ht]
\begin{lstlisting}[style=prompt]
# Modified system instruction:
You are an expert legal reasoning model generating persona-conditioned COUNTERFACTUAL TEXT REWRITES for a legal information retrieval dataset.

Your task is to produce {augmentation_count} COUNTERFACTUAL TEXT REWRITES, each written from the perspective of a different TARGET PERSONA. [...]

# New section inserted after task description:
### Persona Rules

You will produce one rewritten text per persona listed below. Each persona must have a clearly unique tone, rhetorical style, and framing.

PERSONAS:
{persona_dict}

For each persona P:
- Write ONLY in the style of persona P.
- Avoid copying tone, rhetorical patterns, or stylistic decisions from any other persona.
- Ensure strong stylistic separation across personas.
- Still preserve Essentials (E) exactly.

# Modified closing instruction:
Produce {augmentation_count} COUNTERFACTUAL TEXT REWRITES, one for each persona, ensuring strong stylistic diversity across personas while preserving Essentials (E) exactly.
\end{lstlisting}
\caption{Persona-specific extensions to the base prompt. Comments (lines starting with \#) indicate where each modification is applied.}
\label{fig:persona-extensions}
\end{figure*}

\subsection{Persona Descriptions}\label{app:persona-desc}

We define ten legal personas, each characterized by distinct voice, orientation, style,
and constraints. Detailed descriptions follow.

\paragraph{Appellate Judge (Majority).}
Voice/Tone: formal, authoritative, doctrinally precise.
Orientation: emphasizes legal standards, precedent coherence, and institutional stability.
Style: neutral judicial prose, structured reasoning, succinct case citations.
Key features: frames the rule clearly; applies the legal test systematically; expresses conclusions with institutional confidence; avoids emotional or argumentative rhetoric.

\paragraph{Appellate Judge (Dissenting).}
Voice/Tone: assertive, critical, more rhetorical than majority.
Orientation: challenges the majority's application of law; stresses fairness or doctrinal risk.
Style: sharper transitions, explicit disagreement, highlights consequences of the rule.
Key features: points out flaws in reasoning; stresses competing precedent or alternative interpretations; uses expressive but still judicial language.

\paragraph{Prosecutor.}
Voice/Tone: assertive, confident, enforcement-focused.
Orientation: public safety, rule-of-law, strong interpretation of statutes and precedent.
Style: highlights factual elements that justify government action; emphasizes culpability or legitimacy of state conduct.
Key features: stresses why legal standards support the government's position; uses persuasive tone within legal boundaries; underscores societal interest or procedural integrity.

\paragraph{Defense Attorney.}
Voice/Tone: protective, rights-centered, adversarial where necessary.
Orientation: fairness, burden of proof, procedural safeguards, statutory protection.
Style: emphasizes mitigating facts, narrow readings of precedent, constitutional concerns.
Key features: highlights overreach, improper inference, or government burden; frames facts in a defendant-favorable way; uses careful legal language.

\paragraph{Law Professor.}
Voice/Tone: analytical, conceptual, explanatory.
Orientation: doctrine, theory, policy implications.
Style: abstract reasoning, comparative references to broader jurisprudence.
Key features: explains legal standards in a teaching tone; frames issues in terms of doctrinal evolution; uses academic transitions (``conceptually,'' ``doctrinally,'' ``historically'').

\paragraph{Trial Judge.}
Voice/Tone: pragmatic, procedural, fact-sensitive.
Orientation: case management, evidentiary sufficiency, application of law to record facts.
Style: grounded judicial prose, attentive to procedural posture and standards of review.
Key features: begins from the concrete facts or procedural setting; emphasizes admissibility, burdens, trial-level reasoning; avoids broad doctrinal exposition unless necessary.

\paragraph{Concurring Judge.}
Voice/Tone: formal, reflective, analytically distinct.
Orientation: agrees with the outcome but through different reasoning.
Style: judicial prose that reframes or narrows the doctrinal basis.
Key features: explicitly aligns with the judgment, not necessarily the reasoning; highlights alternative legal rationale or limiting principles; maintains institutional tone without dissenting rhetoric.

\paragraph{Public Defender.}
Voice/Tone: empathetic, rights-focused, institutionally critical.
Orientation: systemic fairness, inequality of resources, constitutional protection.
Style: advocacy-oriented but restrained; foregrounds procedural justice.
Key features: emphasizes power imbalance and due process; frames legal standards defensively and narrowly; stresses safeguards against overreach.

\paragraph{Legal Realist Scholar.}
Voice/Tone: skeptical, analytical, outcome-aware.
Orientation: practical effects, judicial behavior, law-in-action.
Style: academic but critical; de-emphasizes formal doctrine in favor of consequences.
Key features: questions how rules operate in practice; highlights incentives, discretion, and institutional behavior; reframes issues in functional rather than formal terms.

\paragraph{Judicial Clerk.}
Voice/Tone: neutral, precise, synthesis-oriented.
Orientation: issue-spotting, clarity, internal consistency.
Style: clean, structured summaries; balanced presentation of arguments.
Key features: reframes the question crisply; organizes issues logically; avoids advocacy or rhetorical flourish.

\subsection{Persona Combinations}\label{app:persona-comb}

Table~\ref{tab:persona-combinations} shows which personas are included in each set size used in our ablation study.

\begin{table}[ht]
  \centering
  \caption{Persona inclusion by set size.}
  \label{tab:persona-combinations}
  \begin{tabularx}{\linewidth}{@{}>{\raggedright\arraybackslash}Xcccc@{}}
    \toprule
    Persona & Set 3 & Set 5 & Set 7 & Set 10 \\
    \midrule
    Defense Attorney & \checkmark & \checkmark & \checkmark & \checkmark \\
    Prosecutor & \checkmark & \checkmark & \checkmark & \checkmark \\
    Appellate Judge (Majority) & \checkmark & \checkmark & \checkmark & \checkmark \\
    Appellate Judge (Dissenting) & & \checkmark & \checkmark & \checkmark \\
    Law Professor & & \checkmark & \checkmark & \checkmark \\
    Trial Judge & & & \checkmark & \checkmark \\
    Public Defender & & & \checkmark & \checkmark \\
    Legal Realist Scholar & & & & \checkmark \\
    Judicial Clerk & & & & \checkmark \\
    Concurring Judge & & & & \checkmark \\
    \bottomrule
  \end{tabularx}
\end{table}

\section{Fine-tuning Configurations}\label{app:config}

Table~\ref{tab:training-samples} summarizes the number of training samples per configuration. Table~\ref{tab:training-and-chunking-config} presents the hyperparameters used for fine-tuning and the chunking configurations for each retrieval model.

\begin{table}[ht]
  \centering
  \caption{Number of training samples per configuration.}
  \label{tab:training-samples}
  \begin{tabular}{@{}lcc@{}}
      \toprule
      Configuration & CLERC & COLIEE \\
      \midrule
      Original     & 1,000 & $\sim$5,000 (chunked) \\
      Vanilla only & 5,000 & 5,000 \\
      Vanilla mix  & 5,000 & 5,000 \\
      Persona only & 5,000 & 5,000 \\
      Persona mix  & 5,000 & 5,000 \\
      \bottomrule
  \end{tabular}
\end{table}

\begin{table}[ht]
  \centering
  \caption{Training hyperparameters and chunking configurations.}
  \label{tab:training-and-chunking-config}
  \begin{tabular}{@{}llr@{}}
    \toprule
    \textbf{Category} & \textbf{Parameter} & \textbf{Value} \\
    \midrule
    \multirow{5}{*}{\textit{Training}}
      & Seed & 42 \\
      & Epochs & 3 \\
      & Learning rate & 2e-5 \\
      & Warmup ratio & 0.1 \\
      & Temperature & 0.05 \\
    \midrule
    \multirow{5}{*}{\textit{Chunking}}
      & BGE-base & 512 / 80 \\
      & BGE-m3 & 8192 / 512 \\
      & E5-base & 512 / 80 \\
      & LegalBERT & 512 / 80 \\
      & DPR & 512 / 80 \\
    \bottomrule
  \end{tabular}
  \vspace{2pt}

  {\footnotesize \textit{Note:} Chunking values are shown as chunk size / overlap (in tokens).}
\end{table}

\section{Retrieval Results}\label{app:retrieval-results}

Table~\ref{tab:finetune-retrievers} presents the complete recall scores for all fine-tuned retrievers on the COLIEE and CLERC datasets.

\begin{table*}[ht]
    \centering
    \small
    \setlength{\tabcolsep}{4pt}
    \renewcommand{\arraystretch}{1.15}
    \begin{tabular}{@{}llcccccccc@{}}
        \toprule
        \textbf{Model} & \textbf{Approach} &
        \multicolumn{4}{c}{\textbf{COLIEE}} &
        \multicolumn{4}{c}{\textbf{CLERC}} \\
        \cmidrule(lr){3-6} \cmidrule(lr){7-10}
        & & R@1 & R@5 & R@10 & R@20
        & R@1 & R@5 & R@10 & R@20 \\
        \midrule
        \textbf{BGE-base-en-v1.5} & Baseline & \textbf{0.076} & 0.228 & 0.316 & 0.413 & 0.586 & 0.782 & 0.826 & 0.880 \\
         & Original & \underline{0.075} & 0.247 & 0.361 & 0.470 & \textbf{0.718} & \underline{0.900} & 0.924 & \underline{0.962} \\
         & Persona Only & \underline{0.075} & \textbf{0.263} & \textbf{0.375} & \underline{0.497} & 0.688 & 0.880 & 0.916 & 0.952 \\
         & Persona Mix & \underline{0.075} & \textbf{0.263} & \textbf{0.375} & \textbf{0.498} & \textbf{0.718} & \textbf{0.902} & \underline{0.926} & \textbf{0.964} \\
         & Vanilla Only & 0.074 & 0.255 & 0.364 & 0.475 & 0.702 & 0.896 & \underline{0.926} & \underline{0.962} \\
         & Vanilla Mix & 0.074 & \underline{0.256} & \underline{0.372} & 0.477 & \underline{0.716} & 0.898 & \textbf{0.928} & \underline{0.962} \\
        \midrule
         \textbf{DPR} & Baseline & 0.058 & 0.189 & 0.264 & 0.330 & 0.548 & 0.762 & 0.822 & 0.874 \\
          & Original & 0.064 & 0.214 & 0.322 & \underline{0.431} & \underline{0.604} & \underline{0.820} & \underline{0.876} & 0.906 \\
          & Persona Only & \textbf{0.072} & \underline{0.226} & \textbf{0.337} & \textbf{0.439} & \textbf{0.614} & 0.808 & \textbf{0.880} & \textbf{0.924} \\
          & Persona Mix & \underline{0.070} & \underline{0.226} & \textbf{0.337} & \textbf{0.439} & 0.594 & 0.796 & 0.866 & 0.916 \\
          & Vanilla Only & 0.063 & \textbf{0.228} & \underline{0.323} & 0.428 & \underline{0.604} & \textbf{0.826} & \textbf{0.880} & 0.916 \\
          & Vanilla Mix & \underline{0.070} & 0.216 & 0.319 & 0.429 & 0.596 & 0.808 & 0.870 & \underline{0.918} \\
        \midrule
        \textbf{LegalBERT} & Baseline & 0.028 & 0.083 & 0.125 & 0.184 & 0.322 & 0.500 & 0.588 & 0.680 \\
         & Original & \textbf{0.062} & \textbf{0.207} & \textbf{0.282} & \textbf{0.378} & \textbf{0.488} & \textbf{0.696} & \textbf{0.778} & \textbf{0.846} \\
         & Persona Only & 0.044 & 0.143 & 0.211 & 0.288 & 0.458 & 0.674 & 0.738 & 0.804 \\
         & Persona Mix & \underline{0.061} & \underline{0.174} & \underline{0.255} & \underline{0.339} & 0.394 & 0.594 & 0.676 & 0.740 \\
         & Vanilla Only & 0.041 & 0.127 & 0.166 & 0.230 & 0.442 & 0.664 & 0.742 & 0.808 \\
         & Vanilla Mix & 0.040 & 0.124 & 0.171 & 0.236 & \underline{0.478} & \underline{0.678} & \underline{0.760} & \underline{0.828} \\
        \midrule
        \textbf{E5-base} & Baseline & 0.077 & 0.235 & 0.317 & 0.405 & 0.688 & 0.842 & 0.894 & 0.932 \\
         & Original & 0.091 & 0.274 & 0.401 & 0.523 & \textbf{0.732} & \underline{0.902} & \underline{0.928} & \underline{0.952} \\
         & Persona Only & \underline{0.092} & 0.278 & 0.413 & \underline{0.543} & 0.710 & 0.892 & 0.924 & \textbf{0.954} \\
         & Persona Mix & \textbf{0.093} & 0.278 & 0.412 & 0.542 & \underline{0.724} & 0.894 & 0.922 & 0.946 \\
         & Vanilla Only & \underline{0.092} & \underline{0.287} & \textbf{0.419} & \underline{0.543} & 0.722 & \textbf{0.906} & \textbf{0.930} & \underline{0.952} \\
         & Vanilla Mix & \underline{0.092} & \textbf{0.293} & \underline{0.417} & \textbf{0.544} & 0.718 & 0.898 & 0.922 & 0.944 \\
        \midrule
        \textbf{BGE-M3} & Baseline & 0.044 & 0.155 & 0.244 & 0.324 & 0.744 & 0.892 & 0.920 & 0.946 \\
         & Original & \textbf{0.064} & \underline{0.217} & 0.301 & \textbf{0.428} & \textbf{0.806} & \underline{0.934} & 0.948 & 0.956 \\
         & Persona Only & 0.057 & 0.216 & \underline{0.310} & 0.412 & \underline{0.800} & \underline{0.934} & 0.950 & \underline{0.962} \\
         & Persona Mix & 0.055 & 0.212 & \textbf{0.318} & 0.419 & \underline{0.800} & 0.932 & \underline{0.954} & \textbf{0.966} \\
         & Vanilla Only & \underline{0.063} & \textbf{0.218} & 0.297 & 0.409 & 0.794 & 0.932 & 0.950 & 0.958 \\
         & Vanilla Mix & \underline{0.063} & 0.210 & \underline{0.310} & \underline{0.423} & 0.798 & \textbf{0.938} & \textbf{0.956} & \underline{0.962} \\
        \bottomrule
    \end{tabular}
    \caption{Recall scores of fine-tuned retrievers on COLIEE and CLERC. Best scores are in \textbf{bold}, and second-best scores are \underline{underlined}.}
    \label{tab:finetune-retrievers}
\end{table*}

\end{document}